\pdfoutput=1

\documentclass[11pt]{article}

\usepackage{naacl2021}

\usepackage{times}
\usepackage{latexsym}

\usepackage[T1]{fontenc}
\usepackage[utf8]{inputenc}
\usepackage{microtype}

\usepackage{graphicx}
\usepackage{caption,subcaption}
\usepackage{amsmath}
\usepackage{booktabs}
\usepackage{multirow}

\newcommand{\comment}[3]{\ensuretext{\textcolor{#3}{[#1 #2]}}}

\newcommand{\com}[1]{}

\newcommand{\marker}[1]{#1:}

\newcommand{\michael}[1]{\comment{\marker{MICHAEL}}{#1}{mdgreen}}
\newcommand{\roy}[1]{\comment{\marker{ROY}}{#1}{orange}}
\newcommand{\yonatan}[1]{\comment{\marker{YONATAN}}{#1}{red}}
\newcommand{\resolved}[1]{}

\title{Automatic Generation of Contrast Sets from Scene Graphs:\\Probing the Compositional Consistency of GQA}

\newcommand{\authorspace}[0]{\quad}

\author{
  Yonatan Bitton$^{\diamondsuit}$ \authorspace 
 Gabriel Stanovsky$^{\diamondsuit}$
 \authorspace 
 Roy Schwartz$^{\diamondsuit}$
 \authorspace
 Michael Elhadad$^{\spadesuit}$\\
 $^{\diamondsuit}$School of Computer Science and Engineering, The Hebrew University of Jerusalem,  Jerusalem, Israel \\
 $^{\spadesuit}$ Department of Computer Science, Ben Gurion University, 
   Beer Sheva, Israel \\
  \{yonatanbitton,gabis,roys\}@cs.huji.ac.il \authorspace elhadad@cs.bgu.ac.il 
}

\begin{document}

\maketitle

\begin{abstract}

Recent works have shown that supervised models often exploit data artifacts to achieve good test scores while their performance severely degrades on samples outside their training distribution.
\textit{Contrast sets} \cite{gardner-etal-2020-evaluating} quantify this phenomenon by perturbing test samples in a minimal way such that the output label is modified.  
While most contrast sets were created manually, requiring intensive annotation effort, we present a novel method which leverages rich semantic input representation to \emph{automatically} generate contrast sets for the visual question answering task. Our method computes the answer of perturbed questions, thus vastly reducing annotation cost and enabling thorough evaluation of models' performance on various semantic aspects (e.g., spatial or relational reasoning). We demonstrate the effectiveness of our approach on the popular GQA dataset \cite{hudson2019gqa} and its semantic scene graph image representation.
We find that, despite GQA's compositionality and carefully balanced label distribution, two strong models drop 13--17\%\resolved{\roy{use a long hyphen (-- and not -) when using numeric ranges}} in accuracy on our automatically-constructed contrast set compared to the original validation set.
Finally, we show that our method can be applied to the \emph{training} set to mitigate the degradation in performance, opening the door to more robust models.\footnote{Our contrast sets and code are available at \url{https://github.com/yonatanbitton/AutoGenOfContrastSetsFromSceneGraphs}.}
\end{abstract}

\section{Introduction}
\begin{figure}[!ht]
\centering
\newcommand{\figlen}[0]{\columnwidth}
    \includegraphics[width=\figlen]{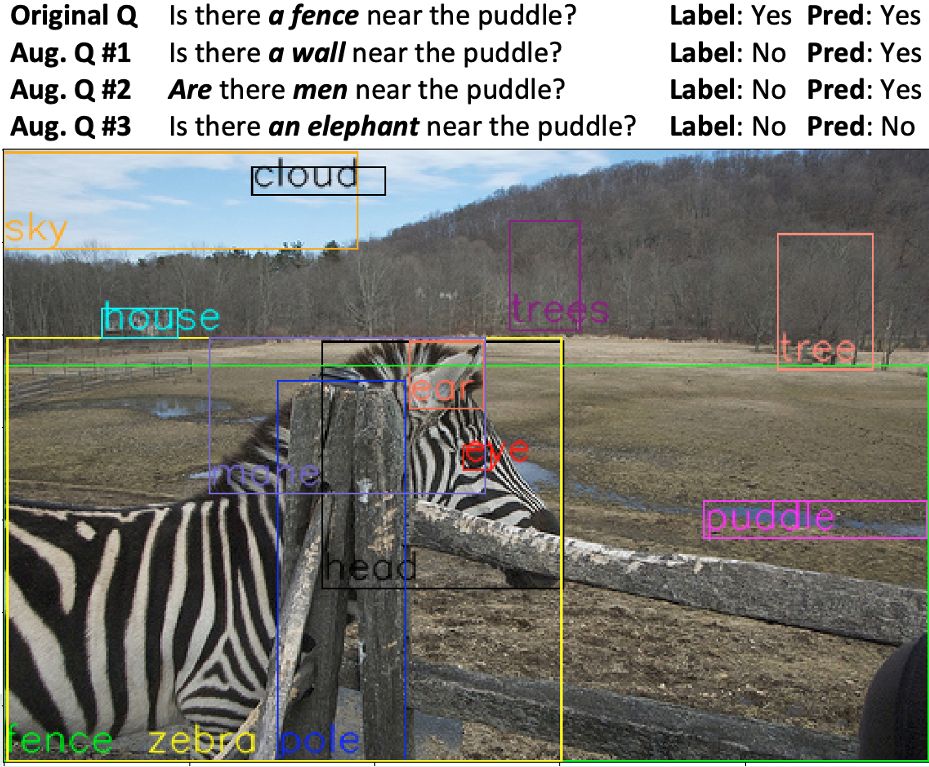}\\%
    \includegraphics[width=\figlen]{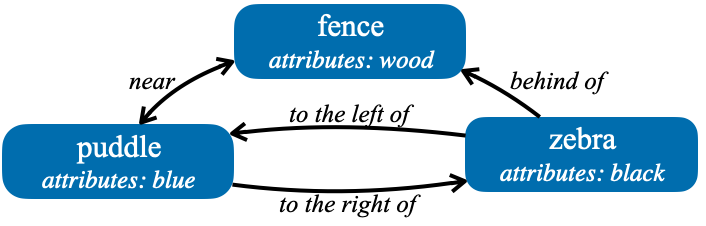}

  \caption{
  Illustration of our approach based on an example from the GQA dataset. Top: QA pairs and an image annotated with bounding boxes from the scene graph. Bottom: relations among the objects in the scene graph. 
  First line at the top is the original QA pair, while  the following 3 lines show our \textit{pertubated} questions: replacing a single element in the question (\textit{a fence}) with other options (\textit{a wall, men, an elephant}), leading to  a change in the output label. 
  For each QA pair, the LXMERT predicted output is shown.
  }
    \label{fig:perturbed-example}
\end{figure}
NLP benchmarks typically evaluate in-distribution generalization, where test sets are drawn \textit{i.i.d} from a distribution similar to the training set. Recent works showed that high performance on test sets sampled in this manner is often achieved by exploiting systematic gaps, annotation artifacts, lexical cues and other heuristics, rather than learning meaningful task-related signal. As a result, the out-of-domain performance of these models is often severely deteriorated~\cite{jia-liang-2017-adversarial,ribeiro-etal-2018-semantically,gururangan-etal-2018-annotation, geva-etal-2019-modeling, mccoy-etal-2019-right, feng-etal-2019-misleading, stanovsky-etal-2019-evaluating}.
Recently, \citet{kaushik2019learning} and \citet{gardner-etal-2020-evaluating} introduced\com{(2) For the NLVR2 VQA variant \cite{suhr2019nlvr2} and other non VQA tasks,} the \textit{contrast sets} approach to probe out-of-domain generalization. Contrast sets are constructed via  minimal modifications to  test inputs, such that their label is modified. For example, in Fig.~\ref{fig:perturbed-example}, replacing ``a fence'' with ``a wall'', changes the answer from ``Yes'' to ``No''. Since such perturbations introduce minimal additional semantic complexity, robust models are expected to perform similarly on the test and contrast sets. However, a range of NLP models\com{including VQA datasets such as NLVR2 \cite{suhr2019nlvr2}, } severely degrade in performance on contrast sets, hinting that they do not generalize well~\cite{gardner-etal-2020-evaluating}. \com{(e.g., the answer to a ``Is there'' question is more often ``yes'' than ``no'' rather than more meaningful task signal)}Except two recent exceptions for textual datasets~\cite{li-etal-2020-linguistically, rosenman-etal-2020-exposing}, contrast sets have so far been built manually, requiring extensive human effort and expertise. 

\com{To mitigate these biases,
several visual question answering (VQA) works constructed \textit{more balanced datasets}: 
most notably, \citet{goyal2017making} designed VQA V2, where the original VQA dataset is balanced such that every question is associated with a pair of similar images that result in two different answers to the question.
\citet{hudson2019gqa}\roy{When a citation is used as an entity in the sentence (
e.g., subject or object), cite it with citet and not cite}addressed the observed lack of generalization by designing GQA, a large-scale dataset of compositional questions over real-world images, starting from images from the Visual Genome, and ``leveraging semantic representations of both the scenes and questions to mitigate language priors and conditional biases and enable fine-grained diagnosis for different question types''.\roy{Is this a quote from the GQA paper? if so, put in italics and cite the paper again.} 
}

In this work, we propose a method for \emph{automatic} generation of large contrast sets for visual question answering (VQA). We experiment with the GQA dataset~\cite{hudson2019gqa}\resolved{\roy{this paper is cited twice, once 2018 and the other 2019}}. 
GQA includes semantic scene graphs~\cite{krishna2017visual} representing the spatial relations between objects in the image, as exemplified in Fig.~\ref{fig:perturbed-example}.
The scene graphs, along with functional programs that represent the questions, are used to \textit{balance} the dataset, thus aiming to mitigate spurious dataset correlations.
We leverage the GQA scene graphs to create contrast sets, by automatically computing the answers to question perturbations, e.g., verifying that there is no wall near the puddle in Fig.~\ref{fig:perturbed-example}.


We create automatic contrast sets for 29K samples or $\approx$22\% of the validation set. We manually verify the correctness of 1,106 of these samples on Mechanical Turk.
Following, we evaluate two leading models, LXMERT \cite{tan-bansal-2019-lxmert} and MAC \cite{hudson2019gqa} on our contrast sets, and find a 13--17\%\resolved{\roy{use -- (see abstract comment)}} reduction in performance \com{(MAC from 64.9 to 51.5, LXMERT from 83.9 to 67.2)}compared to the original validation set. 
Finally, we show that our automatic method for contrast set construction can be used to improve performance by employing it during \emph{training}.
We augment the GQA training set with automatically constructed  \emph{training contrast sets} (adding 80K samples to the existing 943K in GQA), and observe that when trained with it, both LXMERT and MAC improve by about $14\%$ on the contrast sets, while maintaining their original validation performance.



Our key contributions are:
(1) We present an automatic method for creating contrast sets for VQA datasets with structured input representations;
(2) We automatically create contrast sets for GQA, and find that for two strong models, performance on the contrast sets is \resolved{significantly\roy{only use significantly if you computed statistical significance. Use ``substantially'' instead}}lower than on the original validation set; and
(3) We apply our method to augment the training data, improving both models' performance on the contrast sets.

\section{Automatic Contrast Set Construction}
\begin{table*}[]
\resizebox{\textwidth}{!}{%
\begin{tabular}{@{}lll@{}}
\toprule
\textbf{Question template}                              & \textbf{Tested attributes}           & \textbf{Example}                                                                                             \\ \midrule
On which side is the \emph{\textbf{X}}?                 & Relational (left vs.~right)          & On which side is the \textbf{\emph{dishwasher}}? $\rightarrow$ On which side are the \textbf{\emph{dishes}}?                   \\\midrule
What color is the \textbf{\textit{X}}?                  & Color identification                 & What color is the \textbf{\emph{cat}}?$\rightarrow$ What color is the \textbf{\emph{jacket}}?                                  \\\midrule
Do you see \textbf{\textit{X}} or \textbf{\textit{Y}}?  & Compositions                     & Do you see \textbf{\emph{laptops}} or cameras?$\rightarrow$ Do you see \textbf{\emph{headphones}} or cameras?            \\ \midrule
Are there \textbf{\textit{X}} near the \textit{Y}?      & \multirow{3}{*}{Spatial, relational} & Are\resolved{\emph{Are}\roy{why is the word ``Are'' in italics? I see that it changes to ``Is'' in the example, but this is confusing as you use italics to mark variables and this is not a variable}} there any \textbf{\emph{cats}} near the boat? $\rightarrow$ \emph{Is} there any \textbf{\emph{bush}} near the boat? \\
Is the \textit{X} \textbf{\textit{Rel}} the \textit{Y}? &                                      & Is the boy to the \textbf{\emph{right}} of the man? $\rightarrow$ Is the boy to the \textbf{\emph{left}} of the man?           \\
Is the \textbf{\textit{X}} \textit{Rel} the \textit{Y}? &                                      & \resolved{\roy{shouldn't Y be bold and not X in this line?}} Is the \textbf{\emph{boy}} to the right of the man? $\rightarrow$ Is the \textbf{\emph{zebra}} to the right of the man?        \\ \bottomrule
\end{tabular}%
}
\caption{Question templates with original question examples, and generated perturbations modifying the answer. Italic text indicates variables, bold text indicates the perturbed atoms.\resolved{\roy{minor: given that we mark perturbed atoms in bold in the template, can we do the same for the example (i.e., instead of italics?)}}}
\label{tab:question_templates}
\end{table*}
 
\label{sec:construction}
To construct automatic contrast sets for GQA
we first identify a large subset of questions requiring specific reasoning skills~(\S\ref{sec:question_identification}). 
Using the scene graph representation, we perturb each question in a manner which changes its gold answer (\S\ref{sec:pertubation}). Finally, we validate the automatic process via crowd-sourcing (\S\ref{sec:mturk}).

\subsection{Identifying Recurring Patterns in GQA}
\label{sec:question_identification}
The questions in the GQA dataset present a diverse set of modelling challenges, as exemplified in Table~\ref{tab:question_templates}, including object identification and grounding, spatial reasoning and color identification. 
Following the contrast set approach, we create perturbations testing whether models are capable of solving questions which require this skill set, but that diverge from their training distribution.

To achieve this, we identify commonly recurring question templates which specifically require such skills. For example, to answer the question ``\textit{Are there any cats near the boat}?'' a  model needs to identify objects in the image (\textit{cats}, \textit{boat}), link them to the question, and identify their relative position.

We identify six question templates, testing various skills (Table~\ref{tab:question_templates}).
We abstract each question template with a regular expression which identifies the question types as well as the physical objects, their attributes (e.g., colors), and spatial relations. Overall, these regular expressions match 29K questions in the validation set ($\approx$22\%)\resolved{\roy{use $\approx$ rather than $\sim$}}, and 80K questions in the training set ($\approx$8\%).

\subsection{Perturbing Questions with Scene Graphs}
\label{sec:pertubation}
\begin{figure*}
\centering
\begin{minipage}{0.3\textwidth}
\includegraphics[width=\textwidth]{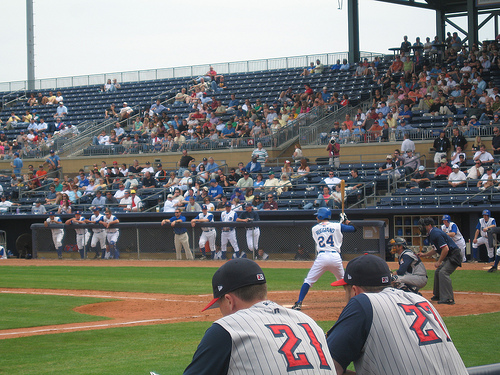}
\end{minipage}
\begin{minipage}{0.45\textwidth}
\centering
\captionsetup{type=table} 
\scalebox{0.65}{
\begin{tabular}{@{}ll@{}}

The \textbf{\textit{bat the batter is holding}} has what color? Brown  $\rightarrow$\\ The \textbf{\textit{helmet}} has what color? Blue                      \\ \midrule
Is there any \textbf{\textit{fence}} near the \textit{players}? Yes $\rightarrow$\\ Are there any \textbf{\textit{trees}} near the \textit{players}? No \\ \midrule
Do you see either \textbf{\textit{bakers}} or \textit{photographers}? No        $\rightarrow$\\ Do you see either \textbf{\textit{spectators}} or \textit{photographers}? Yes   \\ \midrule
Is the \textit{catcher} to the \textit{right} of an \textbf{\textit{umpire}}? No         $\rightarrow$\\ Is the \textit{catcher} to the \textit{right} of a \textbf{\textit{batter}}? Yes         \\ \midrule
Is the \textit{catcher} to the \textbf{\textit{right}} of an \textit{umpire}? No         $\rightarrow$\\ Is the catcher to the \textbf{\textit{left}} of an \textit{umpire}? Yes         \\ 
\end{tabular}
}
\end{minipage}
\caption{GQA image (left) with example perturbations for different question templates (right). Each perturbation aims to change the label in a predetermined manner, e.g., from ``yes'' to ``no''.}
\label{fig:more_examples}
\end{figure*}

We design a perturbation method which guarantees a change in the gold answer for each question template. For example, looking at Fig.~\ref{fig:more_examples}, for the question template \emph{are there \textbf{\textit{X}} near the \textit{Y}?} (e.g., ``Is there any fence near the players?''), we replace either X or Y with a probable distractor (e.g.,\resolved{\roy{no need for italics when writing e.g.}}, replace ``fence'' with ``trees'').

We use the scene graph to ensure that the answer to the question is indeed changed. In our example, this would entail grounding ``players'' in the question to the scene graph (either via exact match or several other heuristics such as hard-coded lists of synonyms or co-hyponyms), locating its neighbors, and verifying that none of them are ``trees.''
We then apply heuristics to fix syntax (e.g., changing from singular to plural determiner, see Appendix~\ref{sec:ling_heuristics}), \resolved{Table~\ref{tab:linguistic_rules}\roy{please point to the appendix section, not the specific table number})} and verify that the perturbed sample does not already exist in GQA.
The specific perturbation is performed per question template. 
In question templates with two objects (\textit{X} and \textit{Y}), we replace \textit{X} with \textit{X'}, such that \textit{X'} is correlated with \textit{Y} in other GQA scene graphs.
In question templates with a single object \textit{X}, we replace \textit{X} with a textually-similar \textit{X'}. For example in the first row in Table~\ref{tab:question_templates} we replace \textit{dishwasher} with \textit{dishes}. Our perturbation code is publicly available.

This process may yield an arbitrarily large number of contrasting samples per question, as there are many candidates for replacing objects participating in questions. We report experiments with up to 1, 3 and 5 contrasting samples per question.

\paragraph{Illustrating the perturbation process.} Looking at Fig.~\ref{fig:perturbed-example}, we see the scene-graph information: \emph{objects} have bounding-boxes around them in the image (e.g., \emph{zebra}); \emph{Objects} have \emph{attributes} (\emph{wood} is an attribute of the \emph{fence} object); and there are \emph{relationships} between the objects (the puddle is to the \textit{right} of the zebra, and it is \emph{near} the fence). The original (question, answer) pair is (``is there  a fence near the puddle?'', ``Yes''). We first identify the question template by regular expressions: ``Is there X near the Y'', and isolate X=\emph{fence}, Y=\emph{puddle}. The answer is ``Yes'', so we know that X is indeed near Y. We then use the existing information given in the scene-graph. We search for X' that is not near Y. To achieve this, we sample a random object (\textit{wall}), and verify that it doesn't exist in the set of scene-graph objects. This results in a perturbed example ``Is there a \emph{wall} near the {puddle}?'', and now the ground truth is computed to be ``No''.\resolved{\michael{Changed `deterministically determined` by compute - I left deterministic in the following paragraph}\roy{A question that reviewers asked and is still unanswered is whether this is done manually or automatically. Can we add a statement such as ``this process is done entirely automatically?'' would that be accurate?}} Consider a different example: (``Is the puddle to the left of the zebra?'', ``Yes''). We identify the question template ``Is the X \textit{Rel} the Y'', where X=\emph{puddle}, Rel=\emph{to the left}, Y=\emph{zebra}. The answer is ``Yes''. Now we can easily change Rel'=\emph{to the right}, resulting in the (question, answer) pair (``Is the puddle to the right of the zebra?'', ``No''). 

We highlight the following: (1) This process is done entirely automatically (we validate it in Section~\ref{sec:mturk}); (2) The answer is deterministic given the information in the scene-graph; (3) We do not produce unanswerable questions. If we couldn't find an alternative atom for which the presuppositions hold, we do not create the perturbed (question, answer) pair; (4) Grounding objects from the question to the scene-graph can be tricky. It can involve exact match, number match (\emph{dogs} in the question, and \emph{dog} in the scene-graph), hyponyms (\emph{animal} in the question, and \emph{dog} in the scene-graph), and synonyms (\emph{motorbike} in the question, and \emph{motorcycle} in the scene-graph). The details are in the published code; (5) The only difference between the original and the perturbed instance is a single atom: an object, relationship, or attribute. 

\subsection{Validating Perturbed Instances}
\label{sec:mturk}
To verify the correctness of our automatic process, we sampled 553 images, each one with an original and perturbed QA pair for a total of 1,106 instances ($\approx$4\% of the validation contrast pairs)\resolved{\roy{$\approx$}}. The (image, question) pairs were answered independently by human annotators on Amazon Mechanical Turk (see Fig.~\ref{fig:amt_example} in Appendix~\ref{app:example}), oblivious to whether the question originated from GQA or from our automatic contrast set.
We found that the workers were able to correctly answer 72.3\% of the perturbed questions, slightly lower than their performance on the original questions (76.6\%).\footnote{The GQA paper reports higher human accuracy (around 90\%) on their original questions. We attribute this difference to the selection of a subset of questions that match our templates, which are potentially more ambiguous than average GQA questions (see Section~\ref{sec:experiments}).} 
We observed high agreement between annotators ($\kappa = 0.679$). 

Our analysis shows that the human performance difference between the perturbed questions and the original questions can be attributed  to the scene graph annotation errors in the GQA dataset: 3.5\% of the 4\% difference is caused by a discrepancy between image and scene graph (objects appearing in the image and not in the graph, and vice versa). Examples are available in Fig.~\ref{fig:scene_graph_mistakes} in Appendix~\ref{sec:examples}.

\begin{table}[t!]
\centering
\scalebox{0.90}{
\begin{tabular}{@{}lrcc@{}}
\toprule
\textbf{Model}            & \textbf{Training set}       & \textbf{Original} & \textbf{Augmented} \\ \midrule
\multirow{2}{*}{MAC}      & Baseline                        & 64.9\%            & 51.5\%             \\
                          & Augmented     & 64.4\%            & 68.4\%             \\ \midrule
\multirow{2}{*}{LXMERT}   & Baseline                        & 83.9\%            & 67.2\%             \\
                          & Augmented    & 82.6\%            & 77.2\%             \\
                          \bottomrule
\end{tabular}
}
\caption{Model accuracy on the original validation set and on our generated contrast sets with maximum of 5 augmentations.
\textit{Baseline} refers to the original models, \textit{augmented} refers to the models trained with our augmented  training contrast sets.\resolved{\roy{is this with 5 augmentations? 1? please add here and on the referring text}}}
\label{tab:training_improvement}
\end{table}

\section{Experiments}
\label{sec:experiments}

We experiment with two top-performing GQA models, MAC~\cite{arad2018compositional} and LXMERT~\cite{tan-bansal-2019-lxmert},\footnote{MAC and LXMERT are the top two models in the \href{https://eval.ai/web/challenges/challenge-page/225/leaderboard/733}{GQA leaderboard} with a public implementation as of the time of submission: \url{https://github.com/airsplay/lxmert} and  \url{https://github.com/stanfordnlp/mac-network/}. \resolved{\roy{Please add a date, or a statement such as. this will most probably not be true forever, and perhaps not even true now}\yonatan{"As of the time of submission" added. Is it ok?}}} to test  their generalization on our automatic contrast sets, leading to various key observations.


\paragraph{Models struggle with our contrast set.}
Table~\ref{tab:training_improvement} 
shows that despite GQA's emphasis on dataset balance and compositionality, both MAC and LXMERT degraded on the contrast set: MAC 64.9\% $\rightarrow$ 51.5\% and LXMERT 83.9\% $\rightarrow$ 67.2\%, compared to only 4\% degradation in human performance.
Full breakdown of the results by template is shown in  Table~\ref{tab:results_single_augmentation}. 
As expected, question templates that reference two objects ($X$ and $Y$) result in larger performance drop compared to those containing a single object ($X$).
Questions about colors had the \emph{smallest} performance drop, potentially because the models performance on such multi-class, subjective questions is relatively low to begin with.

\paragraph{Training on perturbed set leads to more robust models.}
Previous works tried to mitigate spurious datasets biases by explicitly balancing labels during dataset construction \cite{goyal2017making,zhu2016visual7w, zhang2016yin} or using adversarial filtering \cite{zellers-etal-2018-swag,zellers-etal-2019-hellaswag}.
In this work we take an inoculation approach \cite{liu-etal-2019-inoculation} and augment the original GQA training set with contrast training data, resulting in a total of 1,023,607 training samples.
We retrain both models on the augmented training data, and observe in Table~\ref{tab:training_improvement}  that their performance on the \emph{contrast set} almost matches that of the original validation set, with no loss (MAC) or only minor loss (LXMERT) to original validation accuracy.\footnote{To verify that this is not the result of training on more data, we repeated this experiment, removing the same amount of original training instances (so the final dataset size is the same as the original one), and observed very similar results.}
These results indicate that the perturbed \emph{training} set is a valuable signal, which helps models recognize more patterns.  

\begin{table}[tb!]
\centering
\resizebox{1\columnwidth}{!}{%
\begin{tabular}{@{}lrrrr@{}}
\toprule
                                                        & \multicolumn{2}{c}{\textbf{MAC}}                                               & \multicolumn{2}{c}{\textbf{LXMERT}}                                            \\ \midrule
                                                        & \multicolumn{1}{c}{\textbf{Original}} & \multicolumn{1}{c}{\textbf{Aug.}} & \multicolumn{1}{c}{\textbf{Original}} & \multicolumn{1}{c}{\textbf{Aug.}} \\
\text{On which side is the \textbf{\textit{X}}?}                    & 68\%                                  & 57\%                                   & 94\%                                  & 81\%                                   \\
\text{What color is the \textbf{\textit{X}}?}                        & 49\%                                  & 49\%                                   & 69\%                                  & 62\%                                   \\
\text{Are there \textbf{\textit{X}} near the \textit{Y}?}                    & 85\%                                  & 66\%                                   & 98\%                                  & 79\%                                   \\
\text{Do you see \textbf{\textit{X}} or \textbf{\textit{Y}}?}                         & 88\%                                  & 53\%                                   & 95\%                                  & 65\%                                   \\
\text{Is the \textbf{\textit{X}} \textit{Rel} the \textit{Y}?}        & 85\%                                  & 44\%                                   & 96\%                                  & 69\%                                   \\
\text{Is the \textit{X} \textbf{\textit{Rel}} the \textit{Y}?} & 71\%                                  & 38\%                                   & 93\%                                  & 55\%                                   \\
\text{Overall}                                        & 65\%                                  & 52\%                                   & 84\%                                  & 67\%                                   \\ \bottomrule
\end{tabular}
}
\caption{Model accuracy on the original and augmented validation set by question template for a maximum 5 augmentations per instance. }
\label{tab:results_single_augmentation}
\end{table}

\paragraph{Contrast Consistency.}
Our method can be used to generate many augmented questions by simply sampling more items for replacement (Section~\ref{sec:construction}). This allows us to measure the \textit{contrast consistency} \cite{gardner-etal-2020-evaluating} of our contrast set, defined as the percentage of the contrast sets for which a model’s predictions are correct for all examples in the set (including the original example). For example, in Fig.~\ref{fig:perturbed-example} the set size is 4, and only 2/4 predictions are correct. 
We experiment with 1, 3, and 5 augmentations per question with the LXMERT model trained on the original GQA training set. Our results (Table~\ref{tab:consistency}) show that sampling more objects leads to similar accuracy levels for the LXMERT model, indicating that quality of our contrast sets does not depend on the specific selection of replacements. However, we observe that consistency drops fast as the size of the contrast sets per QA instance grows, indicating that model success on a specific instance does not mean it can generalize robustly to perturbations.

\begin{table}[]
\center
\scalebox{0.8}{
\begin{tabular}{cccc}
\toprule
\textbf{\begin{tabular}[c]{@{}c@{}}Augmentations\\per instance\end{tabular}} & \textbf{Contrast sets} & \textbf{Acc.} & \textbf{Consistency} \\ \hline
1                                                                                                      & 11,263                     & 66\%            & 63.4\%               \\
3                                                                                                      & 23,236                     & 67\%            & 51.1\%               \\
5                                                                                                      & 28,968                       & 67\%              & 46.1\%                 \\ \bottomrule
\end{tabular}
}
\caption{Accuracy and consistency results for the LXMERT model on different contrast set sizes.}
\label{tab:consistency}
\end{table}

\section{Discussion and Conclusion}
Our results suggest that both MAC and LXMERT under-perform when tested out of distribution. 
A remaining question is whether this is due to model architecture or dataset design. 
\citet{bogin2020latent} claim that both of these models are prone to fail on compositional generalization because they do not decompose the problem into smaller sub-tasks. Our results support this claim. 
On the other hand, it is possible that  a different dataset could prevent these models from finding shortcuts. Is there a dataset that can prevent \emph{all} shortcuts? 
Our automatic method for creating contrast sets allows us to ask those questions, while we believe that future work in better training mechanisms, as suggested in \citet{bogin2020latent} and  \citet{jin-etal-2020-visually}, could help in making more robust models. 

We proposed an automatic method for creating contrast sets for VQA datasets that use annotated scene graphs.
We created contrast sets for the GQA dataset, which is designed to be compositional, balanced, and robust against statistical biases. We observed a large performance drop between the original and augmented sets. As our contrast sets can be generated cheaply, we further augmented the GQA training data with additional perturbed questions, and showed that this improves models' performance on the contrast set. Our proposed method can be extended to other VQA datasets.

\section*{Acknowledgements}
We thank the reviewers for the helpful comments and feedback.
We thank the authors of GQA for building the dataset, and the authors of LXMERT and MAC for sharing their code and making it usable. This work was supported in part by the Center for Interdisciplinary Data Science Research at the Hebrew University of Jerusalem, and research gifts from the Allen Institute for AI.

\bibliography{naacl2021}

\begin{thebibliography}{23}
\expandafter\ifx\csname natexlab\endcsname\relax\def\natexlab#1{#1}\fi

\bibitem[{Bogin et~al.(2020)Bogin, Subramanian, Gardner, and
  Berant}]{bogin2020latent}
Ben Bogin, Sanjay Subramanian, Matt Gardner, and Jonathan Berant. 2020.
\newblock Latent compositional representations improve systematic
  generalization in grounded question answering.
\newblock \emph{arXiv preprint arXiv:2007.00266}.

\bibitem[{Feng et~al.(2019)Feng, Wallace, and
  Boyd-Graber}]{feng-etal-2019-misleading}
Shi Feng, Eric Wallace, and Jordan Boyd-Graber. 2019.
\newblock \href {https://doi.org/10.18653/v1/P19-1554} {Misleading failures of
  partial-input baselines}.
\newblock In \emph{Proceedings of the 57th Annual Meeting of the Association
  for Computational Linguistics}, pages 5533--5538, Florence, Italy.
  Association for Computational Linguistics.

\bibitem[{Gardner et~al.(2020)Gardner, Artzi, Basmov, Berant, Bogin, Chen,
  Dasigi, Dua, Elazar, Gottumukkala, Gupta, Hajishirzi, Ilharco, Khashabi, Lin,
  Liu, Liu, Mulcaire, Ning, Singh, Smith, Subramanian, Tsarfaty, Wallace,
  Zhang, and Zhou}]{gardner-etal-2020-evaluating}
Matt Gardner, Yoav Artzi, Victoria Basmov, Jonathan Berant, Ben Bogin, Sihao
  Chen, Pradeep Dasigi, Dheeru Dua, Yanai Elazar, Ananth Gottumukkala, Nitish
  Gupta, Hannaneh Hajishirzi, Gabriel Ilharco, Daniel Khashabi, Kevin Lin,
  Jiangming Liu, Nelson~F. Liu, Phoebe Mulcaire, Qiang Ning, Sameer Singh,
  Noah~A. Smith, Sanjay Subramanian, Reut Tsarfaty, Eric Wallace, Ally Zhang,
  and Ben Zhou. 2020.
\newblock \href {https://doi.org/10.18653/v1/2020.findings-emnlp.117}
  {Evaluating models{'} local decision boundaries via contrast sets}.
\newblock In \emph{Findings of the Association for Computational Linguistics:
  EMNLP 2020}, pages 1307--1323, Online. Association for Computational
  Linguistics.

\bibitem[{Geva et~al.(2019)Geva, Goldberg, and
  Berant}]{geva-etal-2019-modeling}
Mor Geva, Yoav Goldberg, and Jonathan Berant. 2019.
\newblock \href {https://doi.org/10.18653/v1/D19-1107} {Are we modeling the
  task or the annotator? an investigation of annotator bias in natural language
  understanding datasets}.
\newblock In \emph{Proceedings of the 2019 Conference on Empirical Methods in
  Natural Language Processing and the 9th International Joint Conference on
  Natural Language Processing (EMNLP-IJCNLP)}, pages 1161--1166, Hong Kong,
  China. Association for Computational Linguistics.

\bibitem[{Goyal et~al.(2017)Goyal, Khot, Summers-Stay, Batra, and
  Parikh}]{goyal2017making}
Yash Goyal, Tejas Khot, Douglas Summers-Stay, Dhruv Batra, and Devi Parikh.
  2017.
\newblock Making the v in vqa matter: Elevating the role of image understanding
  in visual question answering.
\newblock In \emph{Proceedings of the IEEE Conference on Computer Vision and
  Pattern Recognition}, pages 6904--6913.

\bibitem[{Gururangan et~al.(2018)Gururangan, Swayamdipta, Levy, Schwartz,
  Bowman, and Smith}]{gururangan-etal-2018-annotation}
Suchin Gururangan, Swabha Swayamdipta, Omer Levy, Roy Schwartz, Samuel Bowman,
  and Noah~A. Smith. 2018.
\newblock \href {https://doi.org/10.18653/v1/N18-2017} {Annotation artifacts in
  natural language inference data}.
\newblock In \emph{Proceedings of the 2018 Conference of the North {A}merican
  Chapter of the Association for Computational Linguistics: Human Language
  Technologies, Volume 2 (Short Papers)}, pages 107--112, New Orleans,
  Louisiana. Association for Computational Linguistics.

\bibitem[{Hudson and Manning(2019)}]{hudson2019gqa}
Drew~A. Hudson and Christopher~D. Manning. 2019.
\newblock {GQA}: A new dataset for real-world visual reasoning and
  compositional question answering.
\newblock In \emph{Proceedings of the IEEE/CVF Conference on Computer Vision
  and Pattern Recognition (CVPR)}.

\bibitem[{Hudson and Manning(2018)}]{arad2018compositional}
Drew~Arad Hudson and Christopher~D. Manning. 2018.
\newblock \href {https://openreview.net/forum?id=S1Euwz-Rb} {Compositional
  attention networks for machine reasoning}.
\newblock In \emph{International Conference on Learning Representations}.

\bibitem[{Jia and Liang(2017)}]{jia-liang-2017-adversarial}
Robin Jia and Percy Liang. 2017.
\newblock \href {https://doi.org/10.18653/v1/D17-1215} {Adversarial examples
  for evaluating reading comprehension systems}.
\newblock In \emph{Proceedings of the 2017 Conference on Empirical Methods in
  Natural Language Processing}, pages 2021--2031, Copenhagen, Denmark.
  Association for Computational Linguistics.

\bibitem[{Jin et~al.(2020)Jin, Du, Sadhu, Nevatia, and
  Ren}]{jin-etal-2020-visually}
Xisen Jin, Junyi Du, Arka Sadhu, Ram Nevatia, and Xiang Ren. 2020.
\newblock \href {https://doi.org/10.18653/v1/2020.emnlp-main.158} {Visually
  grounded continual learning of compositional phrases}.
\newblock In \emph{Proceedings of the 2020 Conference on Empirical Methods in
  Natural Language Processing (EMNLP)}, pages 2018--2029, Online. Association
  for Computational Linguistics.

\bibitem[{Kaushik et~al.(2019)Kaushik, Hovy, and Lipton}]{kaushik2019learning}
Divyansh Kaushik, Eduard Hovy, and Zachary Lipton. 2019.
\newblock Learning the difference that makes a difference with
  counterfactually-augmented data.
\newblock In \emph{International Conference on Learning Representations}.

\bibitem[{Krishna et~al.(2017)Krishna, Zhu, Groth, Johnson, Hata, Kravitz,
  Chen, Kalantidis, Li, Shamma et~al.}]{krishna2017visual}
Ranjay Krishna, Yuke Zhu, Oliver Groth, Justin Johnson, Kenji Hata, Joshua
  Kravitz, Stephanie Chen, Yannis Kalantidis, Li-Jia Li, David~A Shamma, et~al.
  2017.
\newblock Visual genome: Connecting language and vision using crowdsourced
  dense image annotations.
\newblock \emph{International journal of computer vision}, 123(1):32--73.

\bibitem[{Li et~al.(2020)Li, Shengshuo, Liu, Wu, Zhou, and
  Steinert-Threlkeld}]{li-etal-2020-linguistically}
Chuanrong Li, Lin Shengshuo, Zeyu Liu, Xinyi Wu, Xuhui Zhou, and Shane
  Steinert-Threlkeld. 2020.
\newblock \href {https://doi.org/10.18653/v1/2020.blackboxnlp-1.12}
  {Linguistically-informed transformations ({LIT}): A method for automatically
  generating contrast sets}.
\newblock In \emph{Proceedings of the Third BlackboxNLP Workshop on Analyzing
  and Interpreting Neural Networks for NLP}, pages 126--135, Online.
  Association for Computational Linguistics.

\bibitem[{Liu et~al.(2019)Liu, Schwartz, and Smith}]{liu-etal-2019-inoculation}
Nelson~F. Liu, Roy Schwartz, and Noah~A. Smith. 2019.
\newblock \href {https://doi.org/10.18653/v1/N19-1225} {Inoculation by
  fine-tuning: A method for analyzing challenge datasets}.
\newblock In \emph{Proceedings of the 2019 Conference of the North {A}merican
  Chapter of the Association for Computational Linguistics: Human Language
  Technologies, Volume 1 (Long and Short Papers)}, pages 2171--2179,
  Minneapolis, Minnesota. Association for Computational Linguistics.

\bibitem[{McCoy et~al.(2019)McCoy, Pavlick, and Linzen}]{mccoy-etal-2019-right}
Tom McCoy, Ellie Pavlick, and Tal Linzen. 2019.
\newblock \href {https://doi.org/10.18653/v1/P19-1334} {Right for the wrong
  reasons: Diagnosing syntactic heuristics in natural language inference}.
\newblock In \emph{Proceedings of the 57th Annual Meeting of the Association
  for Computational Linguistics}, pages 3428--3448, Florence, Italy.
  Association for Computational Linguistics.

\bibitem[{Ribeiro et~al.(2018)Ribeiro, Singh, and
  Guestrin}]{ribeiro-etal-2018-semantically}
Marco~Tulio Ribeiro, Sameer Singh, and Carlos Guestrin. 2018.
\newblock \href {https://doi.org/10.18653/v1/P18-1079} {Semantically equivalent
  adversarial rules for debugging {NLP} models}.
\newblock In \emph{Proceedings of the 56th Annual Meeting of the Association
  for Computational Linguistics (Volume 1: Long Papers)}, pages 856--865,
  Melbourne, Australia. Association for Computational Linguistics.

\bibitem[{Rosenman et~al.(2020)Rosenman, Jacovi, and
  Goldberg}]{rosenman-etal-2020-exposing}
Shachar Rosenman, Alon Jacovi, and Yoav Goldberg. 2020.
\newblock \href {https://doi.org/10.18653/v1/2020.emnlp-main.302} {{E}xposing
  {S}hallow {H}euristics of {R}elation {E}xtraction {M}odels with {C}hallenge
  {D}ata}.
\newblock In \emph{Proceedings of the 2020 Conference on Empirical Methods in
  Natural Language Processing (EMNLP)}, pages 3702--3710, Online. Association
  for Computational Linguistics.

\bibitem[{Stanovsky et~al.(2019)Stanovsky, Smith, and
  Zettlemoyer}]{stanovsky-etal-2019-evaluating}
Gabriel Stanovsky, Noah~A. Smith, and Luke Zettlemoyer. 2019.
\newblock \href {https://doi.org/10.18653/v1/P19-1164} {Evaluating gender bias
  in machine translation}.
\newblock In \emph{Proceedings of the 57th Annual Meeting of the Association
  for Computational Linguistics}, pages 1679--1684, Florence, Italy.
  Association for Computational Linguistics.

\bibitem[{Tan and Bansal(2019)}]{tan-bansal-2019-lxmert}
Hao Tan and Mohit Bansal. 2019.
\newblock \href {https://doi.org/10.18653/v1/D19-1514} {{LXMERT}: Learning
  cross-modality encoder representations from transformers}.
\newblock In \emph{Proceedings of the 2019 Conference on Empirical Methods in
  Natural Language Processing and the 9th International Joint Conference on
  Natural Language Processing (EMNLP-IJCNLP)}, pages 5100--5111, Hong Kong,
  China. Association for Computational Linguistics.

\bibitem[{Zellers et~al.(2018)Zellers, Bisk, Schwartz, and
  Choi}]{zellers-etal-2018-swag}
Rowan Zellers, Yonatan Bisk, Roy Schwartz, and Yejin Choi. 2018.
\newblock \href {https://doi.org/10.18653/v1/D18-1009} {{SWAG}: A large-scale
  adversarial dataset for grounded commonsense inference}.
\newblock In \emph{Proceedings of the 2018 Conference on Empirical Methods in
  Natural Language Processing}, pages 93--104, Brussels, Belgium. Association
  for Computational Linguistics.

\bibitem[{Zellers et~al.(2019)Zellers, Holtzman, Bisk, Farhadi, and
  Choi}]{zellers-etal-2019-hellaswag}
Rowan Zellers, Ari Holtzman, Yonatan Bisk, Ali Farhadi, and Yejin Choi. 2019.
\newblock \href {https://doi.org/10.18653/v1/P19-1472} {{H}ella{S}wag: Can a
  machine really finish your sentence?}
\newblock In \emph{Proceedings of the 57th Annual Meeting of the Association
  for Computational Linguistics}, pages 4791--4800, Florence, Italy.
  Association for Computational Linguistics.

\bibitem[{Zhang et~al.(2016)Zhang, Goyal, Summers-Stay, Batra, and
  Parikh}]{zhang2016yin}
Peng Zhang, Yash Goyal, Douglas Summers-Stay, Dhruv Batra, and Devi Parikh.
  2016.
\newblock Yin and yang: Balancing and answering binary visual questions.
\newblock In \emph{Proceedings of the IEEE Conference on Computer Vision and
  Pattern Recognition}, pages 5014--5022.

\bibitem[{Zhu et~al.(2016)Zhu, Groth, Bernstein, and Fei-Fei}]{zhu2016visual7w}
Yuke Zhu, Oliver Groth, Michael Bernstein, and Li~Fei-Fei. 2016.
\newblock Visual7w: Grounded question answering in images.
\newblock In \emph{Proceedings of the IEEE conference on computer vision and
  pattern recognition}, pages 4995--5004.

\end{thebibliography}
\bibliographystyle{acl_natbib}

\vfill\null
\clearpage
\appendix

\section{Appendix}
\label{sec:appendix}

\section*{Ethical Considerations}

\resolved{\roy{Add some context: we built a dataset, automatically, not requiring manual labor, the goal of our dataset is to improve model robustness against variosu biases. Our dataset is built on the GQA dataset.
Then discuss the annotation part, saying this was the only manual work done on the dataset.}}
We created contrast sets automatically, and verified their correctness via the crowdsourcing annotation of a sample of roughly 1K instances. Section~\ref{sec:mturk} describes the annotation process on Amazon Mechanical Turk. The images and original questions were sampled from the public GQA dataset \cite{hudson2019gqa}, in the English language. Fig.~\ref{fig:amt_example} in Appendix~\ref{app:example} provides example of the annotation task. Overall, the crowdsourcing task resulted in  $\approx$6\ \resolved{roy{approx}} hours of work, which paid an average of 11USD per hour per annotator. \resolved{\roy{do you have an estimate on how much time it took them? i.e., how many hours did it take them to make \$55?}}

\paragraph{Reproducibility}
The augmentations were performed with a MacBook Pro laptop. 
Augmentations for the validation data takes < 1 hour per question template, and for the training data < 3 hours per question template. Overall process, < 24 hours. 

The experiments have been performed with the public implementations of MAC~\cite{arad2018compositional} and LXMERT~\cite{tan-bansal-2019-lxmert}, models: \url{https://github.com/airsplay/lxmert}, \url{https://github.com/stanfordnlp/mac-network/}. 
The configurations were modified to not include the validation set in the training process. The experiments were performed with a Linux virtual machine with a NVIDIA’s Tesla V100 GPU. The training took $\sim$1-2 days in each model. Validation took $\sim$ 30 minutes.

\subsection{Generated Contrast Sets Statistics}

Table~\ref{tab:augmentation_report_validation} reports the basic statistics of automatic contrast sets generation method when applied on the GQA validation dataset.
It shows the overall number of images and QA pairs that matched the 6 question types we identified.
Tables~\ref{tab:detailed_validation_augmentation_report} shows the statistics per question type, indicating how productive each augmentation method is.
Tables~\ref{tab:augmentation_report_train} and \ref{tab:detailed_training_augmentation_report} shows the same statistics for the GQA Training dataset.

\begin{table}[h]
\centering
\begin{tabular}{@{}lrrr@{}}
\toprule
\multicolumn{1}{l}{}           & \multicolumn{3}{c}{\textbf{\# Aug. QA pairs}}                                                           \\ \midrule
                               & \multicolumn{1}{l}{\textbf{Max 1}} & \multicolumn{1}{l}{\textbf{Max 3}} & \multicolumn{1}{l}{\textbf{Max 5}} \\
\textbf{\# Images}             & 10,696                             & 10,696                             & 10,696                             \\
\textbf{\# QA pairs}           & 132,062                            & 132,062                            & 132,062                            \\ \midrule
\textbf{\# Aug. QA pairs} & 12,962                             & 26,189                             & 32,802                             \\
\textbf{\# Aug. images}   & 6,166                              & 6,166                              & 6,166                              \\ \midrule
\textbf{\% Aug. images}   & 57.6\%                             & 57.6\%                             & 57.6\%                             \\
\textbf{\% Aug. QA pairs} & 9.8\%                              & 19.8\%                             & 24.8\%                             \\ \bottomrule
\end{tabular}
\caption{Validation data augmentation statistics}
\label{tab:augmentation_report_validation}
\end{table}

\begin{table}[h]
\centering
\begin{tabular}{@{}lrrr@{}}
\toprule
\multicolumn{1}{c}{\multirow{2}{*}{\textbf{Question   template}}} & \multicolumn{3}{c}{\textbf{\# Aug. QA pairs}} \\ \cmidrule(l){2-4} 
\multicolumn{1}{c}{}                                              & \textbf{Max 1}  & \textbf{Max 3}  & \textbf{Max 5} \\ \cmidrule(r){1-1}
On which side is the \textbf{\textit{X}}?                                       & 2,516           & 4,889           & 5,617          \\
What color is the \textbf{\textit{X}}?                                           & 4,608           & 10,424          & 12,414         \\
Are there \textbf{\textit{X}} near the \textit{Y}?                                       & 382             & 867             & 1,320          \\
Do you see \textbf{\textit{X}} or \textbf{\textit{Y}}?                                            & 1,506           & 4,514           & 7,516          \\
Is the \textbf{\textit{X}} \textit{Rel} the \textit{Y}?                           & 766             & 1,314           & 1,392          \\
Is the \textit{X} \textbf{\textit{Rel}} the \textit{Y}?                    & 1,417           & 1,416           & 1,416          \\
 \bottomrule
\end{tabular}
\caption{Augmentation statistics per question template for the validation data}
\label{tab:detailed_validation_augmentation_report}
\end{table}

\begin{table}[h]
\centering
\begin{tabular}{@{}lr@{}}
\toprule
\textbf{\# Images}             & 72,140  \\ 
\textbf{\# QA pairs}           & 943,000 \\ \midrule
\textbf{\# Aug. QA pairs} & 89,936  \\ 
\textbf{\# Aug. images}   & 43,463  \\ \midrule
\textbf{\% Aug. images}   & 60.2\%  \\ 
\textbf{\% Aug. QA pairs} & 9.5\%   \\ 
\bottomrule
\end{tabular}
\caption{Training data augmentation statistics}
\label{tab:augmentation_report_train}
\end{table}

\begin{table*}[h]
\centering
\scalebox{0.9}{
\begin{tabular}{lrrr}
\toprule
\textbf{Question   template}                   & \textbf{\# Aug. QA pairs} & \textbf{\# Aug. images} & \textbf{\% Aug. questions} \\ \midrule
On which side is the \textbf{\textit{X}}?                    & 17,935                           & 16,224                        & 2.2\%                           \\ 
What color is the \textbf{\textit{X}}?                        & 32,744                          & 27,704                        & 4.1\%                           \\ 
Are there \textbf{\textit{X}} near the \textit{Y}?                    & 2,682                            & 2,323                          & 0.3\%                          \\ 
Do you see \textbf{\textit{X}} or \textbf{\textit{Y}}?                         & 10,666                          & 9,704                        & 1.1\%                           \\ 
Is the \textbf{\textit{X}} \textit{Rel} the \textit{Y}?        & 6,302                            & 5,479                          & 0.6\%                           \\ 
Is the \textit{X} \textbf{\textit{Rel}} the \textit{Y}? & 9,938                          & 8,007                        & 1.1\%                           \\ 
\bottomrule
\end{tabular}
}
\caption{Augmentation statistics per question template for the training data}
\label{tab:detailed_training_augmentation_report}
\end{table*}

\newpage
\subsection{Models Performance Breakdown by Question Type and Number of Augmentations}

Table~\ref{tab:results_single_augmentation} shows the breakdown of the performance of the MAC and LXMERT models per question type, on both the original GQA validation set and on the augmented contrast sets on validation.

The LXMERT model has two stages of training: pre-training on several datasets (which includes GQA training and validation data) and fine-tuning. 
To avoid inflating results on the validation data, 
we re-trained the pre-training stage without the GQA data, and fine-tuned on the training sets.  Table~\ref{tab:training_improvement}. We discovered lower performance on the original set (-$\sim$5\%) with both models, but the same improvement on the augmented set (+$\sim$10).

\begin{table*}[h]
\scalebox{0.70}{
\begin{tabular}{@{}lrrrrrrrrrrr@{}}
\toprule
\multirow{3}{*}{\textbf{}}                              & \multicolumn{3}{c}{\multirow{2}{*}{\textbf{Original Dataset}}} & \multicolumn{8}{c}{\textbf{Aug. dataset}}                                                                                       \\ \cmidrule(l){5-12} 
                                                        & \multicolumn{3}{c}{}                                           & \multicolumn{2}{c}{\textbf{Max 1}} & \multicolumn{3}{c}{\textbf{Max 3}}             & \multicolumn{3}{c}{\textbf{Max 5}}             \\ \cmidrule(lr){2-4}
                                                        & \textbf{Size}       & \textbf{MAC}      & \textbf{LXMERT}      & \textbf{MAC}   & \textbf{LXMERT}   & \textbf{Size} & \textbf{MAC} & \textbf{LXMERT} & \textbf{Size} & \textbf{MAC} & \textbf{LXMERT} \\ \cmidrule(r){1-1}
\text{On which side is the \textbf{\textit{X}}?}                    & 2,538               & 68\%              & 94\%                 & 56\%           & 79\%              & 4,927         & 57\%         & 80\%            & 5,662         & 57\%         & 81\%            \\
\text{What color is the \textbf{\textit{X}}?}                        & 4,654               & 49\%              & 69\%                 & 48\%           & 62\%              & 10,506        & 49\%         & 62\%            & 12,498        & 49\%         & 62\%            \\
\text{Are there \textbf{\textit{X}} near the \textit{Y}?}                    & 382                 & 85\%              & 98\%                 & 72\%           & 84\%              & 867           & 69\%         & 80\%            & 1,320         & 66\%         & 79\%            \\
\text{Do you see \textbf{\textit{X}} or \textbf{\textit{Y}}?}                         & 1,506               & 88\%              & 95\%                 & 53\%           & 63\%              & 4,205         & 53\%         & 64\%            & 6,679         & 53\%         & 65\%            \\
\text{Is the \textbf{\textit{X}} \textit{Rel} the \textit{Y}?}        & 766                 & 85\%              & 96\%                 & 42\%           & 67\%              & 1,314         & 44\%         & 69\%            & 1,392         & 44\%         & 69\%            \\
\text{Is the \textit{X} \textbf{\textit{Rel}} the \textit{Y}?} & 1,417               & 71\%              & 93\%                 & 38\%           & 55\%              & 1,417         & 38\%         & 55\%            & 1,417         & 38\%         & 55\%            \\
\text{Overall}                                        & 11,263              & 65\%              & 84\%                 & 50\%           & 66\%              & 23,236        & 51\%         & 67\%            & 28,968        & 52\%         & 67\%            \\ \bottomrule
\end{tabular}
}
\caption{Model accuracy by question template and maximum number of augmentations. 
\\ Italic text indicates variables, bold text indicates the perturbed atoms.}
\label{tab:results_multiple_augmentation}
\end{table*}

\newpage
\subsection{Linguistic Heuristics for Questions Generation}
\label{sec:ling_heuristics}

For each question type, we select an object in the image scene graph, and update the question by substituting the reference to this object by another object.  
When substituting one object by another, we need to adjust the question to keep it fluent.  Table~\ref{tab:linguistic_rules} shows the specific linguistic rules we verify when performing this substitution.

\begin{table*}[h]
\centering
\scalebox{0.80}{
\begin{tabular}{|c|c|c|}
\hline
\textbf{Linguistic rule}            & \textbf{Explanation}                                                                                                       & \textbf{Examples}                                                                                                                                                                                                                      \\ \hline
Singular vs.~plural       & \begin{tabular}[c]{@{}c@{}}If the noun is singular and countable:\\ add ``a'' or ``an''\\If needed, replace ``Are'' and ``Is''\end{tabular}                  & \begin{tabular}[c]{@{}c@{}}“a fence”, “men”\\ “a boy”, “an elephant”\end{tabular}                                                                                                                                             \\ \hline
Definite vs.~indefinite   & \begin{tabular}[c]{@{}c@{}}Do not change definite articles\\  to indefinite articles, and vice versa\end{tabular} & \begin{tabular}[c]{@{}c@{}}”is there any fence near the boy”\\ suggests that there is a boy in the scene graph,\\ which is not always correct\end{tabular}                                                                    \\ \hline
General vs.~specific      & \begin{tabular}[c]{@{}c@{}}Meaning can be changed\\ When replacing to general\\ or specific terms\end{tabular}    & \begin{tabular}[c]{@{}c@{}}“Cats in the image” =\textgreater “Animals in   the image”,   \\ “Animals not in the image” =\textgreater “cats   not in the image”,\\ The opposite directions not necessarily holds\end{tabular} \\ \hline
Countable vs.~uncountable & \begin{tabular}[c]{@{}c@{}}If the noun is uncountable, \\ do not add “a” or “an”\end{tabular}                      & “A cat”, “water”                                                                                                                                                                                                              \\ \hline
\end{tabular}
}
\caption{Partial linguistic rules to notice using our method.}
\label{tab:linguistic_rules}
\end{table*}

\newpage
\subsection{Annotation Task for Verifying Generated Contrast Sets}
\label{app:example}

\begin{figure}[h]
\centering
\includegraphics[width=0.55\textwidth]{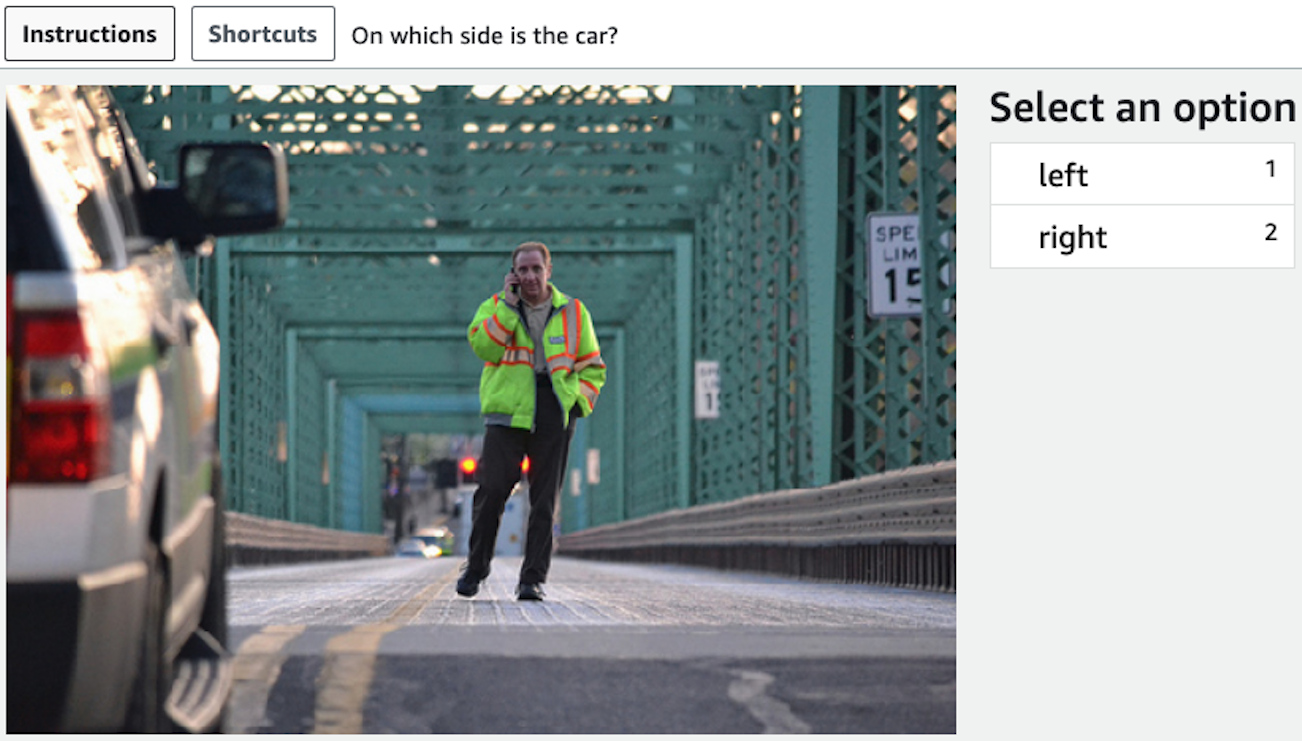}
\caption{Example of the annotation task at the Amazon Mechanical Turk website}
\label{fig:amt_example}
\end{figure}

Fig.~\ref{fig:amt_example} shows the annotation task that is shown to Turkers to validate the QA pairs generated by our method.

\newpage
\subsection{Examples}
\label{sec:examples}

\begin{figure*}[h]
\centering
\includegraphics[width=0.6\textwidth]{images/example_teddy.jpg}
\caption[A table inside a caption]{
\begin{tabular}{@{}|l|l|@{}}
\toprule
\multicolumn{1}{|c|}{\textbf{Original   QA}}         & \multicolumn{1}{c|}{\textbf{Augmented QA}}           \\ \midrule
On which side is the \textbf{\textit{blanket}}? Right                            & On which side is the \textbf{\textit{ornament}}? Left                                                                      \\ \midrule
What color is the \textbf{\textit{teddy bear to the right of the pillow}}? Brown & What color is the \textbf{\textit{christmas lights}}? Yellow                                                               \\ \midrule
Is there a \textbf{\textit{couch}} near the \textit{blanket}? Yes                         & Is there a \textbf{\textit{cat}} near the \textit{blanket}? No                                                                      \\ \midrule
Do you see a \textbf{\textit{pillow}} or \textbf{\textit{couch}} there? Yes                        & Do you see a \textbf{\textit{dress}} or a \textbf{\textit{carpet}} there? No                                                                 \\ \midrule
If the \textit{pillow} to the \textit{left} of a \textbf{\textit{cat}}? No                         & Is the \textit{pillow} to the \textit{left} of a \textbf{\textit{teddy bear}}? Yes                                                           \\ \midrule
Is the \textit{pillow} to the \textbf{\textit{left}} of a \textit{cat}? No                         & No aug. - No relation between (pillow, cat) \\ \bottomrule
\end{tabular}}
\end{figure*}

\begin{figure*}[h]
    \centering 
\begin{subfigure}{0.4\textwidth}
  \includegraphics[width=\linewidth]{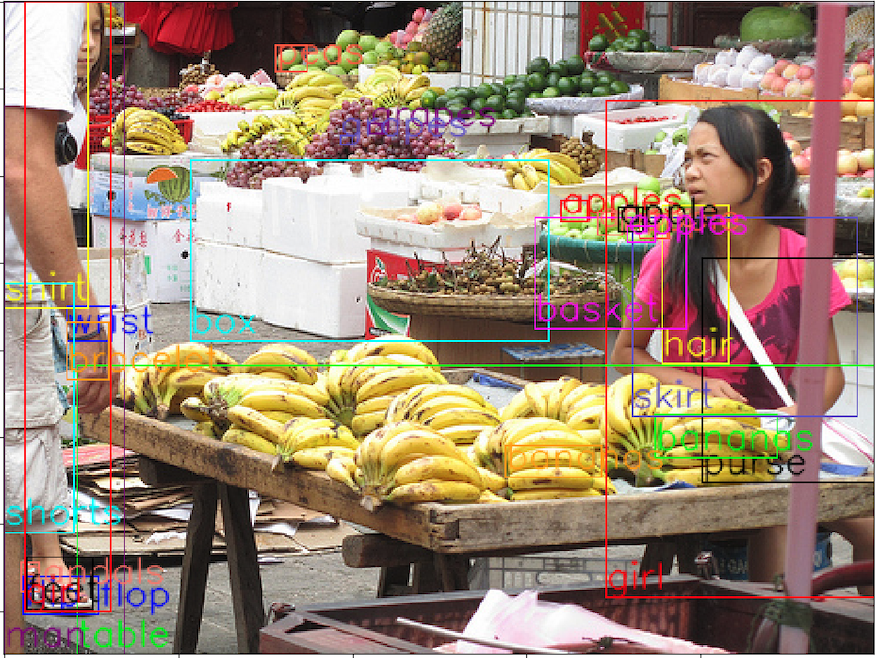}
  \caption{First case example - multiple objects\\
  \textbf{Augmented question}: On which side of the photo are the bananas? \\ \textbf{Expected answer}: right \\  ``bananas'' are annotated in green text color in the right side of the image, but it also appears in additional locations}
\end{subfigure}\hfil 
\begin{subfigure}{0.4\textwidth}
  \includegraphics[width=\linewidth]{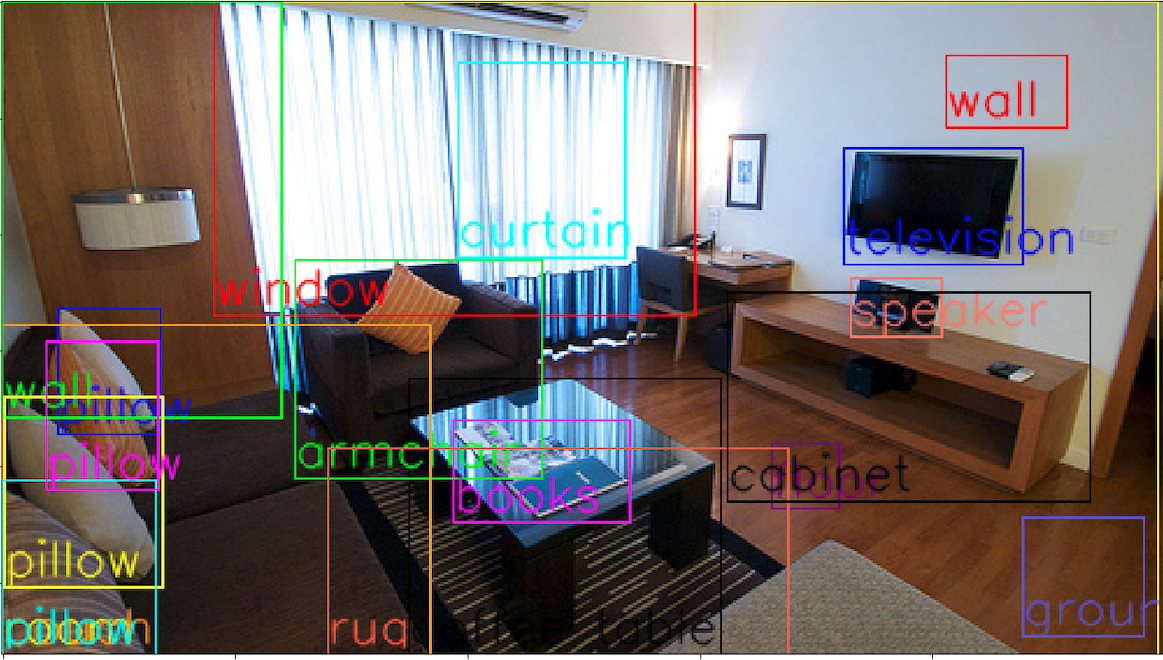}
\caption{Second case example - missing annotation\\
  \textbf{Augmented question}: Do you see either a brown chair or couch in this picture? \\ \textbf{Expected answer}: No \\  We can see a couch in the left side of the image which is not annotated in the scene graph}\end{subfigure}\hfil 
\begin{subfigure}{0.4\textwidth}
  \includegraphics[width=\linewidth]{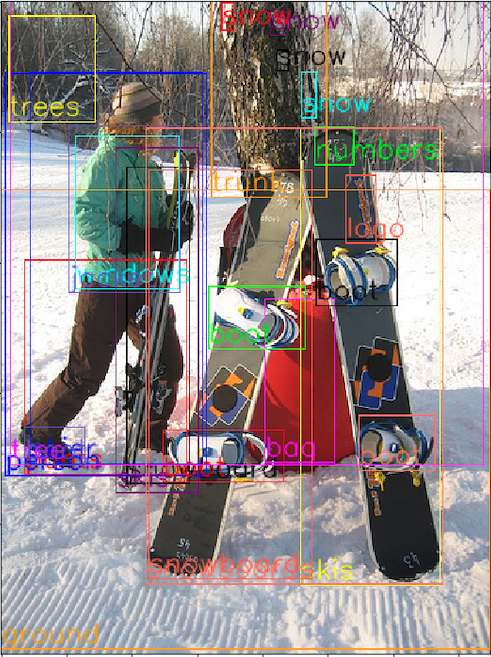}
  \caption{Third case example - incorrect annotation\\
  \textbf{Augmented question}: Do you see either any windows or fences? \\ \textbf{Expected answer}: Yes \\  We can see an incorrect annotation of ``windows'' on the person shirt in azure text color. }
\end{subfigure}\hfil 

\caption{Scene graph annotation mistakes}
\label{fig:scene_graph_mistakes}
\end{figure*}

\end{document}